\documentclass[sigconf]{acmart}

\usepackage{graphicx}
\usepackage{amsmath}
\usepackage{booktabs}
\usepackage{epsfig}
\usepackage{multirow}
\usepackage{algorithm}
\usepackage{algorithmic}
\usepackage{bbm}
\usepackage{colortbl}
\usepackage{caption, subcaption}
\usepackage[utf8]{inputenc}
\usepackage[capitalize]{cleveref}
\usepackage[misc]{ifsym}

\newcommand{\x}{\boldsymbol{x}}

\newcommand{\p}{\boldsymbol{p}}

\newcommand{\R}{\mathbb{R}}

\newcommand{\tablestyle}[2]{\setlength{\tabcolsep}{#1}\renewcommand{\arraystretch}{#2}\centering\footnotesize}
\newcommand\blfootnote[1]{%
 \begingroup
 \renewcommand\thefootnote{}\footnote{#1}%
 \addtocounter{footnote}{-1}%
 \endgroup
}

\newcolumntype{x}[1]{>{\centering\arraybackslash}p{#1pt}}
\newcolumntype{y}[1]{>{\raggedright\arraybackslash}p{#1pt}}
\newcolumntype{z}[1]{>{\raggedleft\arraybackslash}p{#1pt}}

\AtBeginDocument{%
  \providecommand\BibTeX{{%
    \normalfont B\kern-0.5em{\scshape i\kern-0.25em b}\kern-0.8em\TeX}}}

\copyrightyear{2022} 
\acmYear{2022} 
\setcopyright{acmcopyright}\acmConference[MM '22]{Proceedings of the 30th ACM
International Conference on Multimedia}{October 10--14, 2022}{Lisboa, Portugal}
\acmBooktitle{Proceedings of the 30th ACM International Conference on Multimedia
(MM '22), October 10--14, 2022, Lisboa, Portugal}
\acmPrice{15.00}
\acmDOI{10.1145/3503161.3548229}
\acmISBN{978-1-4503-9203-7/22/10}

\acmSubmissionID{1983}

\begin{document}

\title{Making the Best of Both Worlds: A Domain-Oriented \\ Transformer for Unsupervised Domain Adaptation}

\author{Wenxuan Ma}
\affiliation{%
  \institution{Beijing Institute of Technology}
  \state{Beijing}
  \country{China}
}
\email{wenxuanma@bit.edu.cn}

\author{Jinming Zhang}
\affiliation{%
  \institution{Beijing Institute of Technology}
  \state{Beijing}
  \country{China}
}
\email{jinming-zhang@bit.edu.cn}

\author{Shuang Li\textsuperscript{\Letter}}
\affiliation{%
  \institution{Beijing Institute of Technology}
  \state{Beijing}
  \country{China}
}
\email{shuangli@bit.edu.cn}

\author{Chi Harold Liu}
\affiliation{%
  \institution{Beijing Institute of Technology}
  \state{Beijing}
  \country{China}
}
\email{liuchi02@gmail.com}

\author{Yulin Wang}
\affiliation{%
  \institution{Tsinghua University}
  \state{Beijing}
  \country{China}
}
\email{wang-yl19@mails.tsinghua.edu.cn}

\author{Wei Li}
\affiliation{%
  \institution{Inceptio Tech.}
  \state{Shanghai}
  \country{China}
}
\email{liweimcc@gmail.com}

\renewcommand{\shortauthors}{Wenxuan Ma et al.}

\begin{abstract}
Extensive\blfootnote{\Letter~:Corresponding author.} studies on Unsupervised Domain Adaptation (UDA) have propelled the deployment of deep learning from limited experimental datasets into real-world unconstrained domains. Most UDA approaches align features within a common embedding space and apply a shared classifier for target prediction. However, since a perfectly aligned feature space may not exist when the domain discrepancy is large, these methods suffer from two limitations. First, the coercive domain alignment deteriorates target domain discriminability due to lacking target label supervision. Second, the source-supervised classifier is inevitably biased to source data, thus it may underperform in target domain. To alleviate these issues, we propose to simultaneously conduct feature alignment in two individual spaces focusing on different domains, and create for each space a domain-oriented classifier tailored specifically for that domain. Specifically, we design a Domain-Oriented Transformer (DOT) that has two individual classification tokens to learn different domain-oriented representations, and two classifiers to preserve domain-wise discriminability. Theoretical guaranteed contrastive-based alignment and the source-guided pseudo-label refinement strategy are utilized to explore both domain-invariant and specific information. Comprehensive experiments validate that our method achieves state-of-the-art on several benchmarks.
Code is released at \url{https://github.com/BIT-DA/Domain-Oriented-Transformer}.
\vspace{-.75em}

\begin{figure}[!ht]
    \centering
    \includegraphics[width=0.49\textwidth]{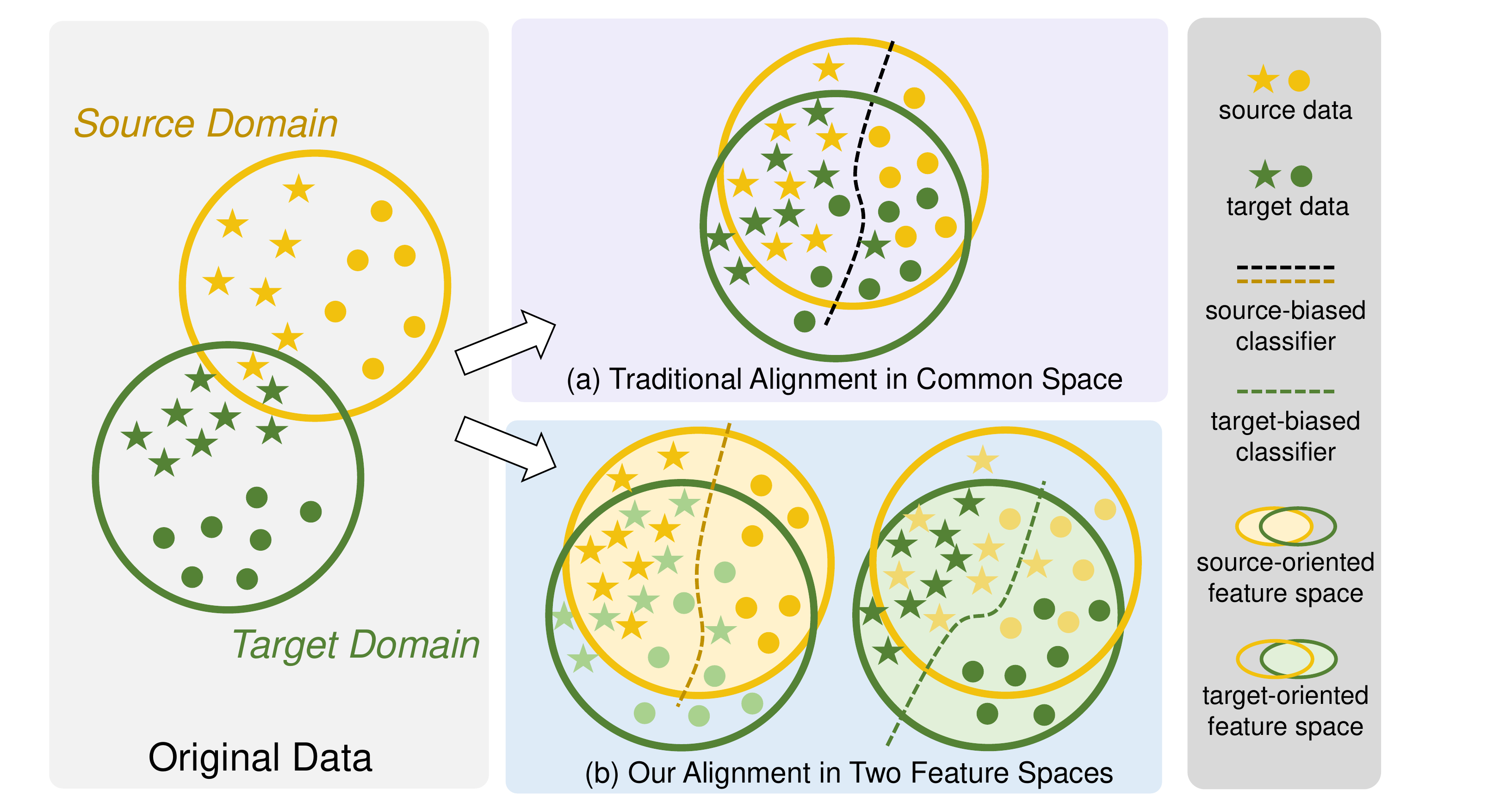}
    \vspace{-2.5em}
    \caption{An illustration to show the major difference between classical UDA paradigm and ours. We develop two domain-oriented feature spaces simultaneously to alleviate the discriminability deterioration problem during imperfect domain alignment.}
    \vspace{-2em}
    \label{Fig_front}
\end{figure}

\end{abstract}



\maketitle

\section{Introduction}\label{sec:introduction}
There is great interest in the community to deploy deep learning to a variety of multimedia applications, such as media-interpretation~\cite{forgione2018implementation,vukotic2016multimodal} and multimodal retrieval~\cite{chu2020multimodal,long2016composite,yang2021deconfounded}. However, deep neural networks heavily rely on large datasets and have inferior generalization ability to data in distinct domains~\cite{DAsurvey_oza2021,DAsurvey_zhao2020}.
These problems limit their real-world utility when the data in the target domain are insufficient or sampled from different distributions. To address these practical issues and enhance the models' generalization performance, unsupervised domain adaptation (UDA) is introduced to transfer knowledge from a labeled source domain to another related but unlabeled target domain with the presence of \textit{domain shift}~\cite{DAsurvey_yang2010,A-distance} in data distribution.

\begin{figure*}[tb]
  \centering
  \includegraphics[width=0.98\textwidth]{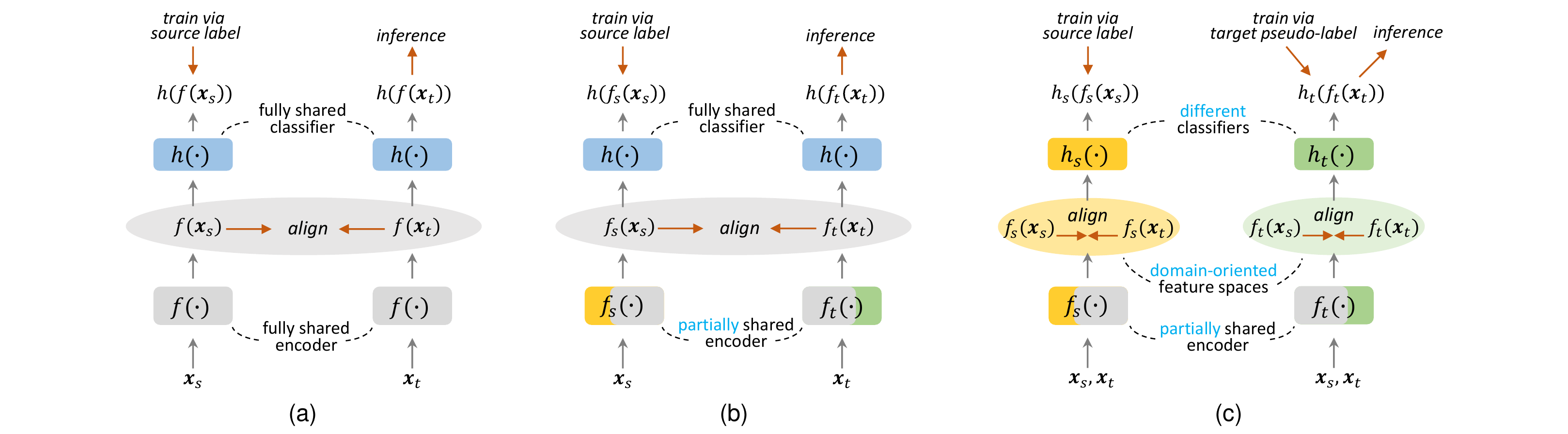}
  \vspace{-1em}
  \caption{Comparison with the existing UDA paradigms to ours. Most UDA methods follow the classical paradigm (a), which adopts a fully-shared encoder as well as a fully-shared classifier. Features from the two domains are aligned to make the source supervised classifier reusable on target. Some works adopt (b), where partially-shared encoders capture domain-specific information additionally. Utilizing the domain specificity, a feature space with better domain-invariance is created to train a shared classifier on it. Different from them, to maximally retain the target-specific characteristics and facilitate accurate target classification, we propose a new architecture (c). It addresses the discriminability degradation problem during single-space domain alignment by leveraging two domain-oriented feature spaces, each maximally benefits its own domain. Discriminative knowledge transfer and exploitation are then achieved through the target pseudo-label assignment in source-oriented feature space, and the final target classifier trained with target supervision.}
  \vspace{-.5em}
  \label{Fig_illstration}
\end{figure*}

Over the past decade, researchers have achieved remarkable improvements on UDA. Following the guidance of the theory~\cite{A-distance} which shows that the error in target domain is primarily bounded by source error and the domain discrepancy, previous works adopt a similar paradigm that \textit{learns to extract domain-invariant representations from a shared encoder and builds one shared classifier based on the invariant feature space}, as shown in Fig.~\ref{Fig_illstration}(a).
Specifically, domain-invariant representations are learned via the domain alignment process, which is either in an explicit way like minimizing the difference in statistic measures across two domains~\cite{DAN,JAN,CORAL} or confusing the domain discriminator~\cite{DANN,CDAN}, or implicitly (e.g., conducting self-training on target data~\cite{CBST,SPCAN,atdoc} or producing target semantic data augmentation~\cite{TSA}).
In the ideal scenario, if features from two domains have achieved perfect class-wise alignment, the classifier supervised by source labels can classify target data correctly.
However, due to lacking true label supervision in the target domain, the class-discriminative information for target domain is easily jeopardized during the domain alignment process~\cite{BSP}, and the shared classifier is inevitably biased to the source domain~\cite{atdoc}, leading to suboptimal decision boundaries for target data. 

Recently, a popular stream of works puts their efforts into incorporating certain domain-specific information to improve the ultimate feature invariance~\cite{DSBN,GDCAN,DWT-MEC}, as illustrated in Fig.~\ref{Fig_illstration}(b). They create domain-specific pathways in the encoder, making it become a partially shared network to better preserve domain-specialized characteristics such as batch statistics~\cite{DSBN} or channel activations~\cite{GDCAN}. These extra features mitigate domain disparity and further improve domain invariance. Although inspiring, the aforementioned methods still follow the paradigm of projecting all the data into a unified domain-invariant feature space. Therefore, when the domain discrepancy is too large for a common space to exist, the same issues such as losing target discriminability and having the classifier source-biased, remain the bottleneck of these approaches. 

As the Hungarian mathematician Cornelius Lanczos used to say ``\textit{The lack of information cannot be filled with any mathematical tricks}'', the loss in target discriminability during domain alignment should be reduced from the beginning. Therefore, we propose a new paradigm that aims to learn two domain-oriented feature spaces with different preference in replace of a single one, and propose two distinctly supervised classifiers in replace of a shared one. Specifically, each pair of feature space and classifier has the goal of maximizing the performance of their respective domain by combining cross-domain invariant knowledge and in-domain intrinsic discriminative information. As illustrated in Fig.~\ref{Fig_illstration}(c), a highly symmetric structure enables target domain to individually possess a maximally discriminative target embedding function and suitable decision boundaries, just as source domain does in previous approaches.

In order to create such two individual feature spaces, we resort to the recent research advances on Vision Transformers (ViTs). The powerful self-attention mechanism enables classification token to adaptively integrate a varying number of image patches, gathering crucial information for recognition~\cite{DoTransSeeLikeConv,evit}. Meanwhile, different class tokens can converge towards dissimilar vectors when trained on different objectives~\cite{DeiT}. Thus, the idea of \textit{one token for one domain} raises naturally, which is to deploy two individual class tokens that respectively learn the source-oriented and target-oriented representations. Moreover, we show in the multiple-source DA experiments that creating distinct class tokens for each domain involved is a straightforward and promising extension to our framework.

To be specific, we propose in this paper a Transformer-based UDA framework dubbed as Domain-Oriented Transformer to simultaneously exploit domain-specific and -invariant information in this new DA paradigm. We adopt two class tokens denoted as [src] and [tgt] to learn different mappings and consequently obtain two different feature spaces. In each space, a domain-specific classifier is trained. To maximally preserve the domain-specific information for creating discriminative feature spaces, we let the two classifiers learn from source labels or target pseudo-labels in their respective domain. Also, we put forward a domain-oriented alignment strategy based on supervised contrastive learning~\cite{supCon} and theoretically show that this objective helps representations in different feature spaces capture correct information from the two original data spaces.
Moreover, to improve the quality of target pseudo-labels and promote knowledge transfer from source to target domain, we propose a new label refinement mechanism originating from our dual classifier architecture. Specifically, target data are divided into reliable and unreliable subsets depending on their performance on the noise-free source classifier, and those in the latter subset are assigned new pseudo-labels according to the reliable ones. We empirically show that this method greatly reduces target label noise via utilizing the structural information inside the source data. Our contributions include:


\begin{itemize}
    \item We propose a new UDA framework that creates two domain-oriented feature spaces for learning different domain-oriented representations. This framework is implemented by creating two classification tokens in a Vision Transformer architecture to integrate different information via self-attention process, together with two individual classifiers.
    \item We propose a domain-oriented alignment objective in each feature space via contrastive learning with theoretical guarantees, as well as a source-guided pseudo-label refinement process to obtain high-quality target pseudo-labels.
    \item The performance of our method on three benchmark UDA datasets surpass both CNN-based and ViT-based UDA methods, especially on the most challenging DomainNet.
\end{itemize}
\section{Related Work}\label{sec:related}
\subsection{Unsupervised Domain Adaptation (UDA)} 
To address the performance degradation issue when deploying models to new environments, UDA is proposed and widely studied~\cite{DAsurvey_zhao2020}, which aims to train a model that can adapt from a labeled source domain to an unlabeled target domain. The mainstream of UDA methods focuses on learning domain-invariant representations with one shared classifier. To align the source and target features, statistical metrics such as maximum mean discrepancy (MMD)~\cite{MMD} between domains are proposed as objectives for models to minimize~\cite{DAN,JAN,DRCN}. Other approaches find inspiration from adversarial training~\cite{GAN} and thus capture domain-invariant representations via a min-max game with the domain discriminator. For example, DANN~\cite{DA_bp} introduces a gradient reversal layer to enable simultaneous optimization of the two players. JADA~\cite{JADA} additionally considers class-wise alignment, GVB~\cite{GVB} adds bridge networks to improve alignment process, and RADA~\cite{RADA} enhances the discriminator using dynamic domain labels. However, alignment without the supervision of true target labels will cause target discriminability deterioration~\cite{BSP}, since target data from different classes might become closer during the process, posing obstacles for the shared classifier to predict target data correctly.

Recently, some works start to pay more attention to domain-specific information learning~\cite{GDCAN,DSBN,DWT-MEC} for an improved alignment. To model domain discriminative representation separately, DSBN~\cite{DSBN} introduces a domain-specific batch normalization mechanism. GDCAN~\cite{GDCAN} designs a domain conditional attention module in convolutional layers to activate distinctly interested channels for each domain, and DWT~\cite{DWT-MEC} makes the domain-specific whitening transform at higher layers. However, these methods still seek to learn a common feature space for a shared classifier, which makes source domain maintain dominant position in training. Different from them, our method exploits the domain-specific information by creating two different feature spaces, letting both domains simultaneously own a more appropriate mapping for better classification.

Our method is also related to another line of research that borrows the pseudo-labeling idea from semi-supervised learning~\cite{pseudo-label}, where the reliable model predictions on unlabeled data are chosen as pseudo-labels to assist the model retraining. Some UDA approaches~\cite{CBST,GMPL,ALDA} adopt target pseudo-labels for a better conditional distribution alignment, such as CBST~\cite{CBST} and MSTN~\cite{MSTN}. To obtain less noisy pseudo-labels, SHOT~\cite{SHOT} and ATDOC~\cite{atdoc} leverage the intrinsic structural information of target data to refine the original labels. All these methods can be regarded as conducting implicit domain alignment by making the source and target features of the same class similar, and they also prevent the classifier from behaving overly source-biased. 
However, these works train both source and target data on the same classifier, which might damage the domain-specific information for both domains. Our method, on the other hand, individually trains two domain-specific classifiers using data and (pseudo-)labels from their respective domains can maximally preserve domain-specific information. Additionally, we propose a source-guided label refinement strategy that to promote knowledge transfer from source to target domain effectively.

\vspace{-.5em}
\subsection{Vision Transformers}
Motivated by the significant improvements of Transformers on natural language processing tasks, researchers apply Transformer architectures to computer vision as a potential alternative to CNN backbones. ViT~\cite{vit} proposes to apply a Transformer encoder with image patches as input to solve image classification problems. Later, DeiT~\cite{DeiT} introduces the distillation token and advanced training strategies that enable ViT to effectively train on much smaller datasets. Recent works such as Swin Transformer~\cite{swin}, PVT~\cite{PVT} and CrossViT~\cite{crossvit} improve the architecture design from different aspects. Furthermore, other researchers apply Vision Transformers to downstream tasks like semantic segmentation~\cite{SETR,volo}, object detection~\cite{DETR,rethinkingDETR} and multimodal tasks~\cite{unit}. 

Since Transformer has the intrinsic advantages to extract more transferable representations, several works~\cite{TransDA,cdtrans} have been proposed to solve domain adaptation with it. For instance, TransDA~\cite{TransDA} injects an attention module after the CNN-based feature extractor to guide the model attention in the source-free UDA setting. TVT~\cite{tvt} applies the domain adversarial training strategy to the classification token as well as the internal transformer blocks. CDTrans~\cite{cdtrans} also adopts the complete transformer architecture and introduces a cross-attention mechanism between the selected source-target pairs. The outputs of the cross-attention branch are then used to supervise the target self-attention branch. 
We take a different perspective from these methods by noticing that the ordinary yet particular classification token in Vision Transformers, via self-attention, are capable of capturing most task-related information~\cite{evit} and learning different mappings, hence are desirable for preserving domain-specific knowledge and learning two differently-oriented feature spaces. Therefore, we propose to train a Transformer with two domain-wise classification tokens, capturing both domain-invariant and specific knowledge for more effective transfer.

\section{Domain-Oriented Transformer}
In this section, we will introduce our framework in detail. Firstly, we provide a brief review of UDA and the self-attention mechanism in Transformers. Then we describe the key design of two domain-wise class tokens as well as the objective functions that help to learn domain-oriented feature spaces. Finally, a new strategy of target pseudo-label refinement based on this framework is presented. 

\begin{figure*}[tb]
  \centering
  \includegraphics[width=0.98\textwidth]{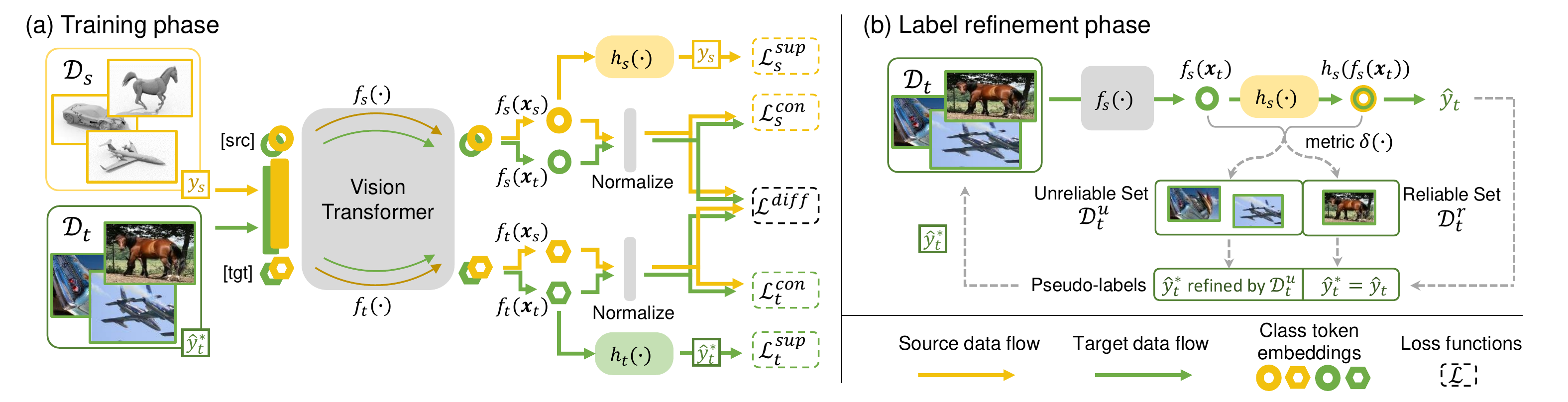}
  \vspace{-1.2em}
  \caption{Illustration of the proposed Domain-oriented Transformer framework. (a) In training phase, two domain-wise class tokens ([src] and [tgt]) are simultaneously sent into the Vision Transformer along with image patches to learn different domain-oriented feature embedding functions $f_s(\cdot),f_t(\cdot)$. Two domain-oriented classifiers and specialized learning objectives using source labels and target pseudo-labels are applied to guide them. (b) In the label refinement phase, we utilize a metric function $\delta(\cdot)$ in source-oriented space to divide the target samples into two individual subsets. Pseudo-labels are reassigned to samples of unreliable subset $\mathcal{T}_u$ according to its cloest reliable centers.}
  \label{Fig_method}
  \vspace{-1em}
\end{figure*}


\vspace{-.5em}
\subsection{Preliminaries}
\paragraph{Unsupervised Domain Adaptation (UDA)}
Contrasting to the assumption of independent and identical distribution, data in UDA is sampled from two different distributions ${P}_s$ and ${P}_t$ to form a labeled source domain $\mathcal{D}_s=\{(\x_{s}^{i},y_{s}^{i})\}_{i=1}^{n_s}$ and an unlabeled target domain $\mathcal{D}_t=\{\x_{t}^{j}\}_{j=1}^{n_t}$, where $n_s$ and $n_t$ denote the number of training data in each domain. A key in this campaign consists in training a model that performs well on the target domain using knowledge transferred from labeled source data. Opposite to most existing methods encouraging domain alignment within a common feature space, this work turns to extracting different domain-oriented embedding spaces for each domain, allowing both domains to have individual projection that maximally preserves class discrinimative knowledge of their own. This is achieved through using the self-attention mechanism within Vision Transformers.
\vspace{-.5em}
\paragraph{Self-attention} The self-attention mechanism is the core of Vision Transformers~\cite{vit}. Given a sequence of input token embeddings $X\in\R^{N\times D}$, three learnable linear projectors $W_Q$, $W_K$, $W_V$ are applied to the layer-normalized features separately to obtain the queries $Q\in\R^{N\times d}$, keys $K\in\R^{N\times d}$ and the values $V\in\R^{N\times d}$. The query sequence is then matched with the keys to get an $N\times N$ self-attention matrix whose elements represent the semantic relevance of the corresponding query-key pairs. Finally, according to the self-attention matrix, new embeddings are calculated in the form of weighted sums over the values:
\begin{equation}\label{equ:attn}
  {\rm Attention}(Q,K,V)={\rm softmax}(\frac{Q K^{\rm T}}{\sqrt d})V.
\end{equation}
Note that self-attention can be viewed as a selective feature aggregation process at each position, using embeddings from other strongly correlated positions. For the class token, it means the most important representations for recognition will be incorporated~\cite{DoTransSeeLikeConv}.

\vspace{-.8em}
\subsection{Learning Domain-Oriented Feature Spaces with Two Domain-Wise Class Tokens}
\paragraph{Supervised Losses on Two Class Tokens}
Since that the classification token in Transformer is designed to adaptively aggregate patch embeddings, it is suitable for learning domain-oriented feature space. Therefore, we use two class tokens to learn source-oriented embedding function $f_s(\cdot)$ and target-oriented embedding function $f_t(\cdot)$ respectively.

Formally, for each input sequence of image features from image patches $X_p = [\x_p^1; \x_p^2; ... ; \x_p^M]\in \R^{M\times D}$ (which can be either source or target), we add two learnable class tokens [src] and [tgt], namely $X = [\x_{[src]};X_p;\x_{[tgt]}]\in \R^{N\times D}$ where $N = M+2$. The corresponding two embeddings output by Transformer encoder are regarded as the source-oriented and target-oriented representation of the input image. Therefore, we can obtain four kinds of representations as $f_s(\x_{s}), \; f_t(\x_{s}), \; f_s(\x_{t}), \; f_t(\x_t)$. Next, we will introduce how to effectively use them for knowledge preservation and transfer.

To begin with, we build a source classifier $h_s(\cdot)$ on $f_s(\cdot)$ and a target classifier $h_t(\cdot)$ on $f_t(\cdot)$. The source labels are then utilized to supervise the source data predictions from source-oriented feature embedding to explore the source-specific information: 

\begin{equation}\label{equ:source_loss}
  \mathcal{L}_{s}^{sup}=\frac{1}{n_s}\sum_{(\x_s,y_s)\in \mathcal{D}_s}{\mathcal{E}\left( h_s(f_s(\x_{s})),{y}_{s}\right)},
\end{equation}
where $\mathcal{E}(\cdot, \cdot)$ denotes the standard cross-entropy (CE) loss.
Similarly, the target classifier and the target-oriented features are supervised only by pseudo-labels in the target domain:
\begin{equation}\label{equ:target_loss}
  \mathcal{L}_{t}^{sup}=\frac{1}{n_t}\sum_{(\x_t,\hat{y}^*_{t})\in \mathcal{D}_t}{\mathcal{E}\left(h_t(f_t(\x_{t})),\hat{y}^*_{t}\right)}.
\end{equation}
Here, $\hat{y}^*_{t}$ is the one-hot pseudo-label for $\x_{t}$. Note that all target training samples are given high-quality pseudo-labels using source knowledge, which we will discuss in \S~\ref{sec:label_refine}.
Additionally, different from ATDOC~\cite{atdoc} which regards a non-parametric model as the auxiliary target classifier to obtain pseudo-labels for training the major classifier, our target classifier is directly trained for target prediction, and it is also used in the final inference stage.

Overall, the supervised loss can be unified as
\begin{equation}\label{equ:sup_loss}
  \mathcal{L}^{sup} = \mathcal{L}_{s}^{sup} + \mathcal{L}_{t}^{sup}.
\end{equation}


\paragraph{Domain-Oriented Contrastive Losses}
The supervised loss maximally explores domain-specific information in each domain, but the classifiers may end up overfitting. Therefore, information from another domain should be leveraged to improve generalization. 
However, as aforementioned, indiscriminate alignment within a shared space potentially impairs the class-wise discriminability preserved, making existing methods suboptimal in leveraging the domain specificity.

To address the issue, we propose a pair of contrastive-based domain alignment losses that, with their deployment in different spaces, are theoretically guaranteed to maintain respective domain-specific information beyond feature alignment.
Specifically, in source-oriented contrastive loss, for a target sample $\x_t^j$ with pseudo-label $\hat{y}^{j*}_{t}$, its positive set consists all source samples with label $y_s=\hat{y}^{*}_{t}$. We denote this positive set as 
$\mathcal{P}_{s,j}=\{\x_s|y_s=\hat{y}^{j*}_{t}\}$. Then, we have source-oriented contrastive loss for $\x_t^j$ as
\begin{equation}\label{equ:source_conloss}
  \mathcal{L}_{s}^{con}(\x_t^j)=\frac{-1}{|\mathcal{P}_{s,j}|} \sum_{\x_s^{+} \in \mathcal{P}_{s,j} }\left [ {\rm log}\frac{{\rm exp}(\tilde{f}_s(\x_t^j)^\top \tilde{f}_s(\x_s^{+})/\tau)}{\sum_{\x_s \in \mathcal{D}_s}{\rm exp}(\tilde{f}_s(\x_t^j)^\top \tilde{f}_s(\x_s)/\tau)}\right ],
\end{equation}
where $\tilde{f}_s(\x)=\frac{f_s(\x)}{\Vert f_s(\x) \Vert}$, and the temperature hyper-parameter $\tau$ controls the concentration level of the distribution~\cite{instdisc}. The total loss $\mathcal{L}^{con}_s$ is the summation over all target samples.

The symmetric form of Eq.~\eqref{equ:source_conloss} is adopted in target-oriented space, where $\mathcal{P}_{t,i}=\{\x_t|\hat{y}^{*}_{t}=y_s^i\}$ represents the positive set for $\x_s$, and
\begin{equation}\label{equ:target_conloss}
  \mathcal{L}_{t}^{con}(\x_s^i)=\frac{-1}{|\mathcal{P}_{t,i}|} \sum_{\x_t^{+} \in \mathcal{P}_{t,i} }\left [ {\rm log}\frac{{\rm exp}(\tilde{f}_t(\x_s^i)^\top \tilde{f}_t(\x_t^{+})/\tau)}{\sum_{\x_t \in \mathcal{D}_t}{\rm exp}(\tilde{f}_t(\x_s^i)^\top \tilde{f}_t(\x_t)/\tau)}\right ],
\end{equation}
The target-oriented contrastive loss is summed over the source samples. We combine this pair of losses to obtain the following unified domain alignment objective
\begin{equation}\label{equ:con_loss}
  \mathcal{L}^{con} = \mathcal{L}_{s}^{con} + \mathcal{L}_{t}^{con}.
\end{equation}
Unlike some recent work employing other types of contrastive loss~\cite{TCL}, the contrastive loss we proposed is directly related to the mutual information which we seek to maximize.

\vspace{.8em}
\noindent \textbf{Remark} \textit{The following two inequalities reveal the relationship between domain-oriented contrastive losses and two specific mutual information. For simplicity, we denote the random variable for representations $f_s(\x_d)$, $f_t(\x_d)$ as $f^d_s$, $f^d_t$, $d\in\{s,t\}$, and have:}

\begin{equation}
  \begin{aligned}
     - \mathcal{L}_{s}^{con} &\le I(f^s_s;f^t_s,y) + \log(N-1) \\
                                      &= I(f^s_s;f^t_s) + I(f^s_s;y|f^t_s) + \log(N-1)
  \end{aligned}
\end{equation}

\begin{equation}
  \begin{aligned}
   - \mathcal{L}_{t}^{con} &\le I(f^t_t;f^s_t,y) +\log(N-1) \\
                          &= I(f^t_t;f^s_t) + I(f^t_t;y|f^s_t) + \log(N-1)
  \end{aligned}
\end{equation}
\textit{Above inequalities show that both losses encourages feature alignment between source and target. Moreover, minimizing source-oriented contrastive loss also maximizes source-specific information for representation $f_s(\cdot)$, while minimizing target-oriented contrastive loss, on the other hand, maximizes target-specific information for representation $f_t(\cdot)$. We give the proofs in the supplementary.}

\paragraph{Regularization on Representation Difference}
The above losses are expected to provide different domain-specific knowledge for respective domains. Therefore, two learned domain-oriented representations for the same input should be distinctive. To explicitly promote this and accelerate the training, we add a regularization term to penalize the similar pair of representations. In this way, source-oriented features space explores more source-specific information and becomes more beneficial for source classification, while the target-oriented space is the opposite, making the best of both domains. This regularization term takes the following form:

\begin{equation}\label{equ:diff_loss}
  \mathcal{L}^{diff} = \frac{1}{n_s+n_t} \sum_{\x \in \mathcal{D}_s \cup \mathcal{D}_t} \left(\frac{\tilde{f}_s(\x)^\top \tilde{f}_t(\x)}{\Vert \tilde{f}_s(\x)\Vert \Vert \tilde{f}_t(\x) \Vert} \right)^2
\end{equation}

\subsection{Target Pseudo-label Refinement}\label{sec:label_refine}
Similar to self-training based UDA methods~\cite{SPCAN,SHOT}, our objective functions also rely on the quality of target pseudo-labels. 
In previous methods, once the shared classifier is trained by noisy pseudo-labels, it may remember its false label and produce overconfident predictions~\cite{overconfident}, making the subsequent refinement hard to improve.
To better exploit the advantage of our symmetric structure, we propose a new label refinement strategy to reduce the noise from original predictions. Our insight here is that the source-oriented feature space and classifier, not being supervised by target pseudo-labels, are less affected by the previous noisy pseudo-labels. Accordingly, they can serve as fair judges to determine whether a target sample is currently well-aligned or not. Hereafter, by evaluating target samples in the source-oriented feature space, we are capable of largely reducing overconfident errors.

To be precise, we evaluate all target samples by a metric $\delta(\cdot)$ and accordingly divide them into reliable target subset $\mathcal{D}_t^r$ and unreliable target subset $\mathcal{D}_t^u$. On one hand, samples in the former set are considered to be easy to transfer or well aligned with source domain, thus we trust their predictions and use them directly as pseudo-labels. On the other hand, we reassign pseudo-labels for samples in the unreliable set according to their nearest reliable center in order to keep the pseudo-labels uniform within the neighborhood.
During the training procedure, we iteratively conduct label refinement until the network converges.

In summary, the procedure of target pseudo-label refinement is as follows. First, the initial predictions for target samples on source-specific classifier are
\begin{equation}\label{equ:pl}
  \hat{y}_t = \mathop{\arg\max}\limits_{k=1,...,K} \p^{(k)} = \mathop{\arg\max}\limits_{k=1,...,K} \sigma^{(k)} \!\left(h_s(f_s(\x_t))\right),
\end{equation}
where $\sigma^{(k)}(\boldsymbol{a}) = \frac{\exp(a^k)}{\sum_j \exp(a^j)}$ and $K$ is the number of categories.
Next, we denote $\x_{t,k}$ to be all the target samples with prediction $\hat{y}_t=k$ and construct class-wise reliable set $\mathcal{D}_t^{r}=\bigcup_k \mathcal{D}_{t,k}^{r}$, where
$\mathcal{D}_{t,k}^{r} = \{\x_t | \x_t \in \x_{t,k},\delta(\x_t) \geq \overline{\delta(\x_{t,k})}\}$ while the rest go to unreliable set $\mathcal{D}_{t}^{u}$.
\newcommand{\tabincell}[2]{\begin{tabular}{@{}#1@{}}#2\end{tabular}}  
\begin{table*}[htbp]
  \centering
  \tiny
  \caption{Accuracy (\%) on DomainNet for unsupervised domain adaption. The column-wise domains are selected as the source domain and the row-wise domains as the target domain. CNN-based methods are presented in the supplementary.}
  \vspace{-8pt} 
   \resizebox{\textwidth}{!}{
  \setlength{\tabcolsep}{0.5mm}{
    \begin{tabular}{|c|ccccccc|c|ccccccc||c|ccccccc|c|ccccccc|}
    \specialrule{.1em}{.05em}{.05em}
    ViT-S~\cite{vit} & {  clp}   & {  inf}   & {  pnt}   & {  qdr}   & {  rel}   & {  skt}   & Avg.  &
    CDTrans-S~\cite{cdtrans} & {  clp}   & {  inf}   & {  pnt}   & {  qdr}   & {  rel}   & {  skt}   & Avg.  & 
    \textbf{DOT-S} & {  clp}   & {  inf}   & {  pnt}   & {  qdr}   & {  rel}   & {  skt}   & Avg. \\
    \hline
    {  clp}  &  --   & 19.3  & 43.2  & 14.3  & 58.8  & 46.4  & 36.4  &  {  clp}  & --    & 24.2  & 47.0  & 22.3  & 64.3  & 50.6  & 41.7  &  {  clp}  & --    & 19.5 &  51.3 &  27.5 &  67.6 &  51.7 &  43.5 \\
    {  inf}  & 35.2  &  --   & 36.7  & 4.7   & 50.4  & 30.0  & 31.4  &  {  inf}  & 45.3  & --    & 45.3  & 6.6   & 62.8  & 38.3  & 39.7  &  {  inf}  & 59.5 & -- &  51.5 &  14.2 &  69.9 &  46.8 &  48.4  \\
    {  pnt}  & 44.7  & 18.7  &  --   & 4.5   & 59.0  & 38.1  & 33.0  &  {  pnt}  & 53.6  & 20.4  & --    & 10.6  & 63.9  & 42.4  & 38.2  &  {  pnt}  & 58.5 &  18.9 & -- &  16.5 &  70.4 &  47.2 &  42.3 \\
    {  qdr}  & 23.2  & 3.3   & 10.1  &  --   & 17.0  & 14.5  & 13.6  &  {  qdr}  & 2.8   & 0.2   & 0.6   & --    & 0.7   & 4.2   & 1.7   &  {  qdr}  & 39.3 &  6.1 &  22.3 & -- &  34.7 &  25.6 &  25.6 \\
    {  rel}  & 48.3  & 18.9  & 50.4  & 7.0   &  --   & 37.0  & 32.3  &  {  rel}  & 47.1  & 17.9  & 45.0  & 7.9   & --    & 31.7  & 29.9  &  {  rel}  & 62.3 &  20.0 &  57.0 &  20.9 & -- &  49.4 &  41.9 \\
    {  skt}  & 54.3  & 16.5  & 41.1  & 15.3  & 53.8  &  --   & 36.2  &  {  skt}  & 61.0  & 19.3  & 46.8  & 22.8  & 59.2  & --    & 41.8  &  {  skt}  & 64.6 &  16.8 &  49.9 &  30.4 &  65.4 & -- &  45.4   \\
    Avg.     & 41.1  & 15.3  & 36.3  & 9.2   & 47.8  & 33.2  & \cellcolor{lightgray}{30.5}  &  Avg.     & 42.0  & 16.4  & 36.9  & 14.0  & 50.2  & 33.4  & \cellcolor{lightgray}{32.2}  &  Avg.     & 56.8 &  16.3 &  46.4 &  21.9 &  61.6 &  44.1 & \cellcolor{lightgray}{\textbf{41.2}} \\
    \specialrule{.1em}{.05em}{.05em}
    \specialrule{.1em}{.05em}{.05em}     ViT-B\cite{vit} & {  clp}   & {  inf}   & {  pnt}   & {  qdr}   & {  rel}   & {  skt}   & Avg.  &     CDTrans-B\cite{cdtrans} & {  clp}   & {  inf}   & {  pnt}   & {  qdr}   & {  rel}   & {  skt}   & Avg.  &     \textbf{DOT-B} & {  clp}   & {  inf}   & {  pnt}   & {  qdr}   & {  rel}   & {  skt}   & Avg. \\     \hline     {  clp}  &   --    & 20.1  & 46.2  & 13.0  & 62.3  & 48.8  & 38.1 & {  clp}  & --    & 27.9  & 57.6  & 27.9  & 73.0  & 58.8  & 49.0  &  {  clp}  & --    & 20.2  & 53.6  & 26.7  & 71.2  & 55.2  & 45.4  \\     {  inf}  &   46.4  & --    & 45.2  & 5.1   & 62.3  & 37.5  & 39.3 & {  inf}  & 58.6  & --    & 53.4  & 9.6   & 71.1  & 47.6  & 48.1  &  {  inf}  & 63.0 &  --   & 54.6  & 12.3  & 73.1  & 50.7  & 50.7    \\     {  pnt}  &   48.1  & 19.1  & --    & 4.4   & 62.5  & 41.8  & 35.2 & {  pnt}  & 60.7  & 24.0  & --    & 13.0  & 69.8  & 49.6  & 43.4  &  {  pnt}  & 61.8  & 20.3   & --    & 11.4  & 72.2  & 50.5  & 43.2  \\     {  qdr}  &   28.2  & 5.2   & 14.4  & --    & 21.9  & 17.7  & 17.5 & {  qdr}  & 2.9   & 0.4   & 0.3   & --    & 0.7   & 4.7   & 1.8  &  {  qdr}  & 47.3  & 7.4  & 30.3   &  --   & 44.6  & 33.7  & 32.7    \\     {  rel}  &   53.2  & 19.3  & 53.5  & 7.2   & --    & 41.6  & 35.0 & {  rel}  & 49.3  & 18.7  & 47.8  & 9.4   & --    & 33.5  & 31.7  &  {  rel}  & 62.9  & 20.0  & 56.9  & 17.3    & --    & 49.3  & 41.3  \\     {  skt}  &   58.0  & 18.5  & 46.5  & 15.7  & 58.7  & --    & 39.5 & {  skt}  & 66.8  & 23.7  & 54.6  & 27.5  & 68.0  & --    & 48.1  &  {  skt}  & 67.3  & 18.7  & 52.9  & 27.8  & 69.8   & --    &  47.3  \\     Avg.     &   46.8  & 16.4  & 41.2  & 9.1   & 53.5  & 37.5  & \cellcolor{lightgray}{34.1} &  Avg.    & 47.7  & 18.9  & 42.7  & 17.5  & 56.5  & 38.8  & \cellcolor{lightgray}{37.0}  &  Avg.     & 60.5  & 17.3  & 49.7  & 19.1  & 66.2  & 47.9  & \cellcolor{lightgray}{\textbf{43.4}}  \\     \specialrule{.1em}{.05em}{.05em}

    \end{tabular}%
    \label{tab:domainnet} }}%
    \vspace{-1.5em}
\end{table*}%
Then, $K$ reliable class centers are calculated, and the refined pseudo-labels are obtained by finding the nearest class center:

\begin{equation}\label{equ:center}
  \boldsymbol{c}_k = \frac{1}{|\mathcal{D}_{t,k}^r|}\sum_{\x_t \in \mathcal{D}_{t,k}^r} f_s(\x_t).
\end{equation}

\begin{equation}\label{equ:label_assignment}
  \hat{y}^*_{t} = \begin{cases}
    \hat{y}_t& \text{if $\x_t \in \mathcal{D}_t^r$}, \\
    \arg \min_{k} w_k \!\cdot\! d(f_s(\x_{t}), \boldsymbol{c}_k)& \text{if $\x_t \in \mathcal{D}_t^u$}.
  \end{cases}
\end{equation}
Here, $d(\cdot, \cdot)$ denotes the distance measure where we use cosine distance, and $w_k=\exp{(\frac{|\mathcal{D}_{t,k}^r|}{|\mathcal{D}_t^{r}|})}$ is a class-wise distance weight related to the size of each class-wise reliable subset. It is based on the consideration that easier-to-transfer categories produce more reliable samples and they also tend to form more compact clusters, therefore samples with equal distance to several class centers are more likely belong to the cluster having less reliable samples.

Finally, we discuss the choices of the metric function $\delta(\cdot)$. As it should return higher values for more reliable samples, we consider confidence score, entropy and negative energy~\cite{energy-ood} as options. The three metrics can be calculated as

\vspace{-1em}
\begin{equation}
  \text{Confidence:} \ \  \delta_{conf}(\x_t) = \max_k \p^{(k)},\quad\quad
\end{equation}
\vspace{-.4em}
\begin{equation}
  \text{Entropy:} \ \quad  \delta_{ent}(\x_t) = -\sum_k \p^{(k)} \log \p^{(k)}, \quad\quad \quad
\end{equation}
\vspace{-.4em}
\begin{equation}
  \quad \ \  \text{Energy:} \ \ \delta_{energy}(\x_t) = \log \sum_k \exp (h_s(f_s(\x_t))^{(k)}).
\end{equation}
Experimentally, $\delta_{energy}$ is the optimal choice. To be clear, we summarize the overall label refinement procedure in Alg.~\ref{alg:label-refinement}.

\vspace{-1em}
\begin{algorithm}[!htbp]
  \small
  \caption{\small Algorithm of Target Pseudo-label Refinement.}
  \label{alg:label-refinement}
  \begin{algorithmic} [1]
    \REQUIRE Source-oriented target features $f_s(\x_t)$; Source-oriented classifier $h_s$; Metric function $\delta(\cdot)$; category number $K$
    \ENSURE Refined target pseudo-labels $\hat{y}^*_{t}$
    \STATE Obtain initial pseudo-labels $\hat{y}_t$ by Eq.~\eqref{equ:pl};
    \FOR{$k=1$ to $K$}
      \STATE Calculate $\overline{\delta(\x_{t,k})}$ by averaging $\delta(\x_t)$ which has $\hat{y}_t=k$;
      \STATE Select $\x_t,\hat{y}_t$ with $\delta(\x_t) \ge \overline{\delta(\x_{t,k})}$ into class-wise reliable set $\mathcal{D}_{t,k}^r$;
      \STATE Obtain reliable class center $c_k$ by Eq.~\eqref{equ:center};
    \ENDFOR
    \STATE Combine all $\mathcal{D}_{t,k}^r$ to $\mathcal{D}_t^r$, put the rest target into unreliable set $\mathcal{D}_t^u$;
    \STATE Obtain refined target pseudo-labels $\hat{y}^*_{t}$ by Eq.~\eqref{equ:label_assignment}.
  \end{algorithmic}
\end{algorithm}

\vspace{-2em}
\subsection{Overall Formulation}
In summary, our objective is the sum of losses with two trade-off parameters $\lambda$ and $\beta$:
\begin{equation}\label{equ:total_loss}
  \mathcal{L}_{DOT} = \mathcal{L}^{sup} + \lambda \mathcal{L}^{con} + \beta \mathcal{L}^{diff}.
\end{equation}
As illustrated in Fig.~\ref{Fig_method}, this highly symmetric architecture learns two domain-oriented feature embedding function and two domain-oriented classifier, enabling a simultaneous exploitation for domain-wise discrimination knowledge and domain-invariant information.

\section{Experiments}

\begin{table*}[htbp]
  \centering
  \Large
  \caption{Accuracy (\%) on VisDA-2017 for unsupervised domain adaption. (IN-1k/21k denotes the pretrained model on ImageNet-1k/21k, and $\sim$ denotes the method having similar amount of parameters to the corresponding backbone.)}
  \label{tab:visda}%
  \vspace{-10pt}
  \resizebox{0.85\textwidth}{!}{%
    \begin{tabular}{l|c|c|cccccccccccc>{\columncolor{lightgray}}c}
    \specialrule{.1em}{.05em}{.05em}
    Method & Pretrained & Params (M) & plane & bcycl & bus & car & horse & knife & mcycl & person &plant & sktbrd & train & truck & Avg. \\
  \hline \hline
 ResNet-101~\cite{resnet} && 44.6 & 55.1 & 53.3 & 61.9 & 59.1 & 80.6 & 17.9 & 79.7 & 31.2 & 81.0 & 26.5 & 73.5 & 8.5 & 52.4 \\
  + JADA~\cite{JADA} &IN-1k& $\sim$ & 91.9 &78.0 & 81.5 & 68.7 & 90.2 & 84.1 & 84.0 & 73.6 & 88.2 & 67.2 & 79.0 & 38.0 & 77.0\\
  + IC$^2$FA~\cite{IC2FA} &IN-1k& $\sim$ & 89.7 & 70.6 & 79.8 & 84.3 & 96.5 & 72.1 & 90.4 & 65.3 & 92.7 & 63.3 & 86.5 & 36.0 & 77.3 \\
  + DSBN~\cite{DSBN} &IN-1k& $\sim$  & 94.7 & 86.7 & 76.0 & 72.0 & 95.2 & 75.1 & 87.9 & 81.3 & 91.1 & 68.9 & 88.3 & 45.5 & 80.2\\

  \hline
  ViT-S~\cite{vit} & & 21.8 & 95.7 & 46.3 & 82.9 & 68.7&  83.4 & 57.1 & \textbf{96.3} & 21.8  & 87.5 & 42.8 & 92.8 & 24.7& 66.7\\
  + \textbf{DOT-S} (\textit{ours})& IN-1k&$\sim$ & 97.8 & 89.7 & 89.8 & 84.9 & 97.2 & 95.8 & 93.2 & 84.6 & 96.3 & 91.1 & 91.3 & 60.0 & 89.3  \\ 
  \hline
  ViT-B && 86.3 & 97.7 & 48.1 & 86.6 & 61.6 & 78.1 & 63.4 & 94.7 & 10.3 & 87.7 & 47.7 & \textbf{94.4} & 35.5 & 67.1\\
  + TVT-B~\cite{tvt}&IN-21k & $\sim$ & 92.9&  85.6 &77.5 & 60.5  &93.6 & 98.2 & 89.4 & 76.4 & 93.6 & 92.0&  91.7 & 55.7 & 83.9\\
  + TVT-B~\cite{tvt}&IN-1k & $\sim$ & --& --& --& --& --& --& --& --& --& --& --& --&85.1\\
  + CDTrans-B~\cite{cdtrans}&IN-1k & $\sim$  & 97.1 & 90.5 & 82.4 & 77.5 & 96.6 & \textbf{96.1} & 93.6 & \textbf{88.6} & \textbf{97.9} & 86.9 & 90.3 & 62.8 & 88.4 \\
  + \textbf{DOT-B} (\textit{ours}) &IN-1k& $\sim$  & \textbf{99.3} & \textbf{92.7} & \textbf{89.0} & \textbf{78.8} & \textbf{98.2} & \textbf{96.1} & 93.1 & 80.2 & 97.6 & \textbf{95.8} & \textbf{94.4} & \textbf{69.0} & \textbf{90.3} \\
    \specialrule{.1em}{.05em}{.05em}
  \end{tabular}
  }%
\end{table*}%

\subsection{Datasets and Setup}
We test and analyze our proposed method on three benchmark datasets in UDA, namely Office-Home~\cite{Office-Home}, VisDA-2017~\cite{VisDA2017} and DomainNet~\cite{DomainNet}. We construct transfer tasks on them following the standard procedure. Detailed descriptions about these datasets and task constructions can be found in the supplementary. Note that unless otherwise specified, all the reported accuracies of target domain come from the target classifier prediction $h_t(f_t(\x_t))$.

\vspace{-1em}
\subsection{Implementation Details}
We adopt ViT-Small (\textbf{S}) and ViT-Base (\textbf{B})~\cite{vit} pretrained on ImageNet-1k~\cite{imagenet} (DeiT~\cite{DeiT} pretrained model) as backbones. We set the base learning rate as 1e-3 on VisDA2017 and DomainNet, while using 3e-4 on Office-Home. Both classifiers are the single fully-connect layer that are randomly initialized and have a 10 times larger learning rate following ~\cite{JADA}. The model is optimized by SGD with momentum 0.9 and weight decay 1e-3. The batch-size is 32 for both domains.
The initial target pseudo-labels are obtained from the source model while we update these pseudo-labels throughout the training. The trade-off parameters $\lambda$ and $\beta$ in Eq.~\eqref{equ:total_loss} are set as $1.0$ and $0.1$ respectively, and the temperature $\tau$ is $0.07$ for all datasets.

\vspace{-1em}
\subsection{Overall Results}
We compare our DOT to various UDA methods using ResNet-50/101 or ViT-S/B as the backbones in the experiments.
We mark the pretrained dataset as well as parameter size for each model.

\begin{table*}[htbp]
  \vspace{-.75em}
    \centering
    \Huge
    \caption{Accuracy (\%) on Office-Home for unsupervised domain adaption.}
    \label{tab:home}%
    \vspace{-10pt}
    \resizebox{0.85\textwidth}{!}{%
      \begin{tabular}{l|c|c|cccccccccccc>{\columncolor{lightgray}}c}
        \specialrule{.1em}{.05em}{.05em}
      Method & Pretrained & Params (M) & Ar$\rightarrow$Cl & Ar$\rightarrow$Pr & Ar$\rightarrow$Re & Cl$\rightarrow$Ar & Cl$\rightarrow$Pr & Cl$\rightarrow$Re & Pr$\rightarrow$Ar & Pr$\rightarrow$Cl &Pr$\rightarrow$Re & Re$\rightarrow$Ar & Re$\rightarrow$Cl & Re$\rightarrow$Pr & Avg. \\
      \hline \hline
      ResNet-50~\cite{resnet}& & 25.6 & 44.9 & 66.3 & 74.3 & 51.8 &61.9 & 63.6 & 52.4 & 39.1 & 71.2 & 63.8 &45.9 & 77.2 & 59.4 \\
      + GDCAN~\cite{GDCAN} &IN-1k & $\sim$ & 57.3 & 75.7 & 83.1 & 68.6 & 73.2 & 77.3 & 66.7 & 56.4 & 82.2 & 74.1 & 60.7 & 83.0 & 71.5\\
      + ATDOC~\cite{atdoc} &IN-1k & $\sim$ & 58.3 & 78.8 & 82.3 & 69.4 & 78.2 & 78.2 & 67.1 & 56.0 & 82.7 & 72.0 & 58.2 & 85.5 & 72.2\\
      + TCL~\cite{TCL}     &IN-1k & $\sim$ & 59.4 & 78.8 & 81.6 & 69.9 & 76.9 & 78.9 & 69.2 & 58.7 & 82.4 & 76.9 & 62.7 & 85.6 & 73.4\\
      
      \hline
      ViT-S~\cite{vit} & & 21.8  & 54.4 & 73.8 & 79.9 & 68.6 & 72.6 & 75.1 & 63.6 & 50.2 & 80.0 & 73.6 & 55.2 & 82.2 & 69.1\\
      + CDTrans-S~\cite{cdtrans} &IN-1k & $\sim$ & 60.6 & 79.5 & 82.4 & 75.6 & 81.0 & 82.3 & 72.5 & 56.7 & 84.4 & 77.0 & 59.1 & 85.5 & 74.7 \\
      + \textbf{DOT-S} (\textit{ours})&IN-1k & $\sim$ & 63.7 & 82.2 & 84.3 & 74.9 & 84.3 & 83.0 & 72.4 & 61.0 & 84.8 & 76.4 & 64.1 & 86.7 & 76.5  \\
      \hline
      ViT-B & & 86.3 & 60.2 & 78.3 & 82.7 & 73.3 & 77.3 & 80.3 & 69.6 & 54.9 & 82.3 & 77.3 & 59.9 & 85.2 & 73.4 \\
      + TVT-B~\cite{tvt} &IN-1k& $\sim$ & --& --& --& --& --& --& --& --& --& --& --& --&78.9\\
      + TVT-B~\cite{tvt} &IN-21k &$\sim$ & \textbf{74.9} & 86.8 & 89.5 & 82.8 & 88.0 & 88.3 & 79.8 & 71.9 & 90.1 & \textbf{85.5} & \textbf{74.6} & 90.6 & 83.6\\
      + CDTrans-B~\cite{cdtrans}&IN-1k & $\sim$ & 68.8 & 85.0 & 86.9 & 81.5 & 87.1 & 87.3 & 79.6 & 63.3 & 88.2 & 82.0 & 66.0 & 90.6 & 80.5 \\
      + \textbf{DOT-B} (\textit{ours})&IN-1k & $\sim$ & 69.0 & 85.6 & 87.0 & 80.0 & 85.2 & 86.4 & 78.2 & 65.4 & 87.9 & 79.7 & 67.3 & 89.3 & 80.1  \\
      + \textbf{DOT-B} (\textit{ours})& IN-21k & $\sim$ & 73.1	 &\textbf{89.1}&	\textbf{90.1}	&\textbf{85.5}&	\textbf{89.4} &	\textbf{89.6} &	\textbf{83.2} &	\textbf{72.1}&	\textbf{90.4}	&84.4	&72.9	&\textbf{91.5}	&\textbf{84.3}\\
      \specialrule{.1em}{.05em}{.05em}
      \end{tabular}
    }%
\end{table*}%

\begin{table*}[htbp]
  \vspace{-.75em}
  \small
  \centering
  \caption{Ablation studies on different components of our method on the Office-Home dataset with ViT-S.}
  \vspace{-10pt}
  \resizebox{1.0\textwidth}{!}{
    \begin{tabular}{l|ccc|cccccccccccc>{\columncolor{lightgray}}c}
    \specialrule{.1em}{.05em}{.05em}
    Method & $\mathcal{L}^{sup}$ & $\mathcal{L}^{con}$ & $\mathcal{L}^{diff}$ & Ar$\rightarrow$Cl & Ar$\rightarrow$Pr & Ar$\rightarrow$Re & Cl$\rightarrow$Ar & Cl$\rightarrow$Pr & Cl$\rightarrow$Re & Pr$\rightarrow$Ar & Pr$\rightarrow$Cl &Pr$\rightarrow$Re & Re$\rightarrow$Ar & Re$\rightarrow$Cl & Re$\rightarrow$Pr & Avg. \\
    \hline
    ViT-S & $\mathcal{L}^{sup}_s$ & \textbf{-} & \textbf{-}   & 54.4 & 73.8 & 79.9 & 68.6 & 72.6 & 75.1 & 63.6 & 50.2 & 80.0 & 73.6 & 55.2 & 82.2 & 69.1 \\
    \hline
    & \checkmark & \textbf{-} & \textbf{-} &  61.3 & 80.2 & 83.0 & 71.4 & 81.9 & 81.8 & 69.8 & 58.4 & 84.9 & 75.7 & 61.4 & 85.9 & 74.6  \\
    \textbf{DOT-S} & \checkmark & \checkmark & \textbf{-} & \textbf{64.2} & \textbf{82.4} & 84.3 & 74.6 & 83.4 & 82.4 & 72.3 & 60.7 & \textbf{85.0} & 76.3 & 63.0 & 86.0 & 76.2  \\
     & \checkmark & \checkmark & \checkmark & 63.7 & 82.2 & \textbf{84.3} & \textbf{74.9} & \textbf{84.3} & \textbf{83.0} & \textbf{72.4} & \textbf{61.0} & 84.8 & \textbf{76.4} & \textbf{64.1} & \textbf{86.7} & \textbf{76.5} \\
     \specialrule{.1em}{.05em}{.05em}
    \end{tabular}
  }\label{tab:ablation}
  \vspace{-1em}
\end{table*}%

\textbf{The results on DomainNet} are shown in Table~\ref{tab:domainnet}, where strong results validate DOT's effectiveness in challenging transfer tasks and unseen target test data. 
We observe that DOT-S outperforms CDTrans on average by $9.0\%$. Out of a total 30 tasks, DOT-S surpasses CDTrans-S on 28 of them by a large margin, especially when the source domain is \textit{Quickdraw}. The significant accuracy boost on these unseen test data proves that our target-oriented classifier, though trained with target samples only, learns to generalize well and will not easily overfit to noisy pseudo-labels.
Moreover, our method achieves satisfying results when transferring to \textit{Real} domain, achieving an average of $61.6\%$ and $66.2\%$ with DOT-S/B.

\textbf{Results on VisDA-2017 and Office-Home} are reported in Table~\ref{tab:visda} and \ref{tab:home}.
We notice that DOT-B achieves a class average accuracy of $90.1\%$ on VisDA-2017, which outperforms all the baseline methods, including DSBN which explores domain-specific information via batch statistics.
On Office-Home, our method also achieves competitive results: DOT (IN-1k) outperforms TVT (IN-1k) by $1.2\%$ on average and DOT (IN-21k) outperforms TVT (IN-21k) by $0.7\%$. These results validate the effectiveness of both our proposed strategies and the Transformer backbone on UDA.

\vspace{-.75em}
\subsection{Insight Analysis}\label{sec:analysis}
In this section, we carry out experiments to fully investigate the influence of each component in our Domain-Oriented Transformer. All the analytical experiments are conducted based on DOT-S.

\vspace{-.5em}
\subsubsection{Ablation Study and Sensitivity Analysis}

To verify that each term in the loss function contributes to our method, we conduct an ablation study as shown in Table~\ref{tab:ablation}. We observe that the performance improves with $\mathcal{L}^{sup}$ exploring the target-specific information, and $\mathcal{L}^{con}$ promoting cross-domain knowledge transfer as well as $\mathcal{L}^{diff}$ increasing the difference between domain-oriented spaces.

\begin{figure*}[tb]
  \centering
  \includegraphics[width=0.95\textwidth]{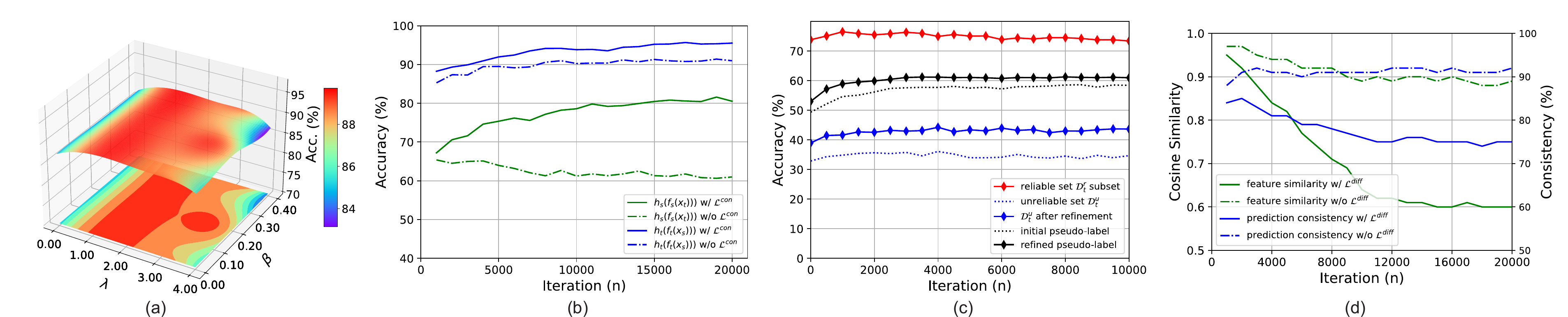}
  \vspace{-1em}
  \caption{Analysis experiments: (a) Parameter sensitivity analysis on Office-Home, (b) Improvements on generalization brought by contrastive-based Alignment on Visda-2017, (c) Development in target pseudo-label accuracy during training process on Office-Home (Pr $\rightarrow$ Cl) (d) Development in difference between class tokens during training process on DomainNet (clp $\rightarrow$ skt).}
  \label{Fig_analysis}
  \vspace{-1em}
\end{figure*}

To show that our method is robust to different choices on hyper-parameters $\lambda$ and $\beta$, we vary their values within a certain range and test their performance on Office-Home using different combinations. The average result is demonstrated in Fig.~\ref{Fig_analysis}(a). 
We see that DOT maintains a stable performance over a wide range of choices.

\begin{table*}[t]
\caption{Comparision results between different variants of DOT on Office-Home dataset. Results are averaged on all 12 tasks.}
\vspace{-1.1em}
\centering
\subfloat[
Comparison of alignment methods.
\label{tab:align}
]{
\centering
\begin{minipage}{0.29\linewidth}{\begin{center}
\tablestyle{4pt}{1.05}
\vspace{-1em}
\begin{tabular}{c|cc}
  \specialrule{.1em}{.05em}{.05em}
    Variant & Method & Acc. \\
    \hline
    \multirow{4}{*}{\begin{tabular}[c]{@{}c@{}}Alignment\\ Methods\end{tabular}} 
    & \multicolumn{1}{l}{w/ MMD~\cite{MMD}}  & 75.5\\
    & \multicolumn{1}{l}{w/ DANN~\cite{DANN}}   & 74.9 \\
    & \multicolumn{1}{l}{w/ TCL~\cite{TCL}}    &  76.3\\
    & \multicolumn{1}{l}{w/ \textit{\textbf{ours}}}    & \textbf{76.5} \\
    \specialrule{.1em}{.05em}{.05em}
  \end{tabular}
\end{center}}\end{minipage}
}
\hspace{1em}
\subfloat[
Comparison of pseudo-labeling methods.
\label{tab:pseudo-labeling}
]{
\begin{minipage}{0.31\linewidth}{\begin{center}
\tablestyle{1pt}{1.05}
\vspace{-1em}
\begin{tabular}{c|cc}
  \specialrule{.1em}{.05em}{.05em}
    Variant & Method & Acc. \\
    \hline
    \multirow{4}{*}{\begin{tabular}[c]{@{}c@{}}Pseudo-labeling\\ Methods\end{tabular}} 
    & \multicolumn{1}{l}{w/ Confidence}  & 72.6 \\
    & \multicolumn{1}{l}{w/ CBST~\cite{CBST}}   & 73.7 \\
    & \multicolumn{1}{l}{w/ SHOT~\cite{SHOT}}  & 76.2 \\
    & \multicolumn{1}{l}{w/ \textit{\textbf{ours}}}    & \textbf{76.5} \\
    \specialrule{.1em}{.05em}{.05em}
  \end{tabular}
\end{center}}\end{minipage}
}
\hspace{1em}
\subfloat[
Comparison of metric function $\delta(\cdot)$ \label{tab:metric}.
]{
\begin{minipage}{0.31\linewidth}{\begin{center}
\tablestyle{6pt}{1.05}
\vspace{-.2em}
\begin{tabular}{c|cccc}
  \specialrule{.1em}{.05em}{.05em}
  Variant & Method & $\mathcal{T}_r$ acc. & $\mathcal{T}_r$ ratio & Acc. \\ 
  \hline
  \multirow{3}{*}{\begin{tabular}[c]{@{}c@{}}Metric\\ Function\end{tabular}} 
  & $\delta_{conf}$ & 85.7 & 59.2 & 75.5 \\
  & $\delta_{ent}$  & 84.8 & 58.3 & 75.4 \\
  & \textbf{$\delta_{energy}$} & 87.5 & 53.6 & \textbf{76.5} \\
  \specialrule{.1em}{.05em}{.05em}
  \end{tabular}
\end{center}}\end{minipage}
}
\vspace{-.1em}
\label{tab:comparisons} 
\end{table*}

\begin{table*}[htbp]
  \vspace{-1em}
  \tiny
  \centering
  \caption{Label refinement within different feature space of our method on the Office-Home dataset with ViT-S.}
  \vspace{-1.5em}
  \resizebox{0.95\textwidth}{!}{
    \begin{tabular}{c|cccccccccccc>{\columncolor{lightgray}}c}
    \specialrule{.1em}{.05em}{.05em}
    Feature space & Ar$\rightarrow$Cl & Ar$\rightarrow$Pr & Ar$\rightarrow$Re & Cl$\rightarrow$Ar & Cl$\rightarrow$Pr & Cl$\rightarrow$Re & Pr$\rightarrow$Ar & Pr$\rightarrow$Cl &Pr$\rightarrow$Re & Re$\rightarrow$Ar & Re$\rightarrow$Cl & Re$\rightarrow$Pr & Avg. \\
    \hline
    Target-oriented $f_t(\x_t)$ & 61.7 & 81.6 & 84.2 & 73.5 & 81.1 & 82.8 & 69.6 & 54.5 & 84.5 & 76.1 & 58.1 & 85.5 & 74.4  \\
    Source-oriented $f_s(\x_t)$ & \textbf{63.7} & \textbf{82.2} & \textbf{84.3} & \textbf{74.9} & \textbf{84.3} & \textbf{83.0} & \textbf{72.4} & \textbf{61.0} & \textbf{84.8} & \textbf{76.4} & \textbf{64.1} & \textbf{86.7} & \textbf{76.5} \\
    \specialrule{.1em}{.05em}{.05em}
    \end{tabular}
  }\label{tab:label_refinement}
  \vspace{-1em}
  \end{table*}%

\vspace{-1em}
\subsubsection{On Contrastive-based Domain Alignment}
\hspace*{\fill} 

\textbf{Comparison of Domain Alignment Methods.}
As shown in Table~\ref{tab:align}, we conduct experiments on different alternatives for the domain alignment loss $\mathcal{L}^{con}$. These alternatives includes:
\textit{Maximum Mean Discrepancy (MMD)}~\cite{MMD}, a classic statistic-based metric, \textit{Domain adversarial training (DANN)}~\cite{DANN} that utilizes the domain discriminator, and \textit{Transferrable Contrastive Loss (TCL)}~\cite{TCL}, another contrastive-based method.
We observe that our contrastive alignment objective outperforms all other variants, since it encourages not only the feature alignment, but also the maximization of domain-specific information that is relevant to the task. 

\textbf{Contrastive-based Alignment Improves Generalization.}
We further investigate the effect of contrastive-based domain alignment in terms of generalization ability. We test target data accuracy on source-specific classifier, (i.e. $h_s(f_s(\x_t))$) and vice versa ($h_t(f_t(\x_s))$). Note that the test data comes from the other domain, and therefore is never trained on the classifier being tested.
As shown in Fig.~\ref{Fig_analysis}(b), after applying contrastive losses (denote as w/ $\mathcal{L}_{con}$), the prediction accuracies on unseen data from the other domain improve significantly, verifying that the alignment process help both feature spaces learn more invariant and robust embeddings. We can also see that without contrastive losses, the accuracy of $h_s(f_s(\x_t))$ decreases, indicating the occurrence of overfitting.

\vspace{-.75em}
\subsubsection{On Pseudo-label Refinement}
\hspace*{\fill} 

\textbf{Comparison to existing pseudo-labeling methods.} Table~\ref{tab:pseudo-labeling} demonstrates the results of DOT-S on Office-Home using different pseudo-labeling methods, including selection-based methods~\cite{CBST} and refinement-based methods~\cite{SHOT}. Note that our strategy can be viewed as a combination of the two. The results prove that our label-refinement strategy works better. Fig.~\ref{Fig_analysis}(c) illustrates how it operates. The reliable subset includes high-quality pseudo-labels, and noisy labels in unreliable subset are reduced after label reassignment, bringing an overall improvement in label quality.


\textbf{Comparison of metrics for selecting reliable target.}
In Table~\ref{tab:metric} we compare different metric functions. We report the average of target pseudo-label accuracy, as well as the sample ratio (i.e. $|\mathcal{D}_t^r|/n_t$) of the reliable set over all 12 tasks during the first selection process. We observe that $\delta_{energy}$ selects the reliable subset with the least amount of noise and helps the model achieve top performance.

\textbf{Which Feature Space for Label Refinement?}
To prove that the label refinement process achieves better results when utilizing source-oriented representation instead of target-oriented ones, we compare the two variants on Office-Home. Specifically, we employ the same label refinement strategy on $f_s(\x_t)$ and $f_t(\x_t)$ respectively and use the obtained pseudo-labels for model training. The results are listed in Table.~\ref{tab:label_refinement}, which indicates that source-oriented representation is more suitable, especially for harder transfer tasks like Pr$\rightarrow$Cl when pseudo-labels before refinement are noisier. We think this improvement is brought by leveraging source knowledge.

\vspace{-.75em}
\subsubsection{On Discrepancy of token embeddings}
Fig.~\ref{Fig_analysis}(d) shows the trend of feature similarity (measured by cosine similarity and prediction consistency using the same classifier $h_t$) with training.
The results validate that the difference between two domain-oriented feature spaces represented by [src] and [tgt] token embeddings increases through training, and $\mathcal{L}^{diff}$ prompts it significantly.

\begin{table}[t]
  \Huge
   \centering
   \caption{Multi-source domain adaptation on Office-Home.}
   \vspace{-8pt}
   \resizebox{0.485\textwidth}{!}{
   \begin{tabular}{l|cccc>{\columncolor{lightgray}}c}
    \specialrule{.1em}{.05em}{.05em}
    Method & Cl,Pr,Re $\rightarrow$ \textbf{Ar} & Ar,Pr,Re $\rightarrow$ \textbf{Cl} & Ar,Cl,Re $\rightarrow$ \textbf{Pr} & Ar,Cl,Pr $\rightarrow$ \textbf{Re} & Avg.\\
    \hline
    ResNet-50 & 49.3 & 46.9 &66.5 &73.6 &59.1 \\
    + DARN~\cite{DARN} & 69.9 & 68.6 & 83.4 &  84.3 &76.5\\
    + WADN~\cite{WADN} & 73.8 & \textbf{70.2} & 86.3 & 87.3  &79.4\\
    \hline
    ViT-S & 75.9 & 59.9 & 82.2 & 83.4& 75.4\\
    + \textbf{DOT-S} (\textit{ours})  &\textbf{79.4} & 65.6 &\textbf{87.1}& \textbf{87.6}& \textbf{79.9}\\
    \specialrule{.1em}{.05em}{.05em}
   \end{tabular}}
   \vspace{-1em}
   \label{tab:msda}
  \end{table}

\subsection{Extension to the Multi-Source Domain Adaptation Setting}
We show that the idea of one class token for one domain can be naturally extended to multiple source setting. Specifically, we create the same amount of domain tokens to the total number of domains. As reported in Table~\ref{tab:msda}, we show that based on a strong performance of ViT-S baseline, DOT outperforms WADN by 0.5\% in average. 
\section{Conclusion}
We propose Domain-Oriented Transformer for UDA. Different from the classical UDA paradigm that learns a domain-invariant representation and a shared classifier, we propose to simultaneously learn two embedding spaces and two classifiers via class tokens in Vision Transformer. Our new paradigm enables the simultaneous exploitation of domain-specific and -invariant information. We propose contrastive alignment losses and source-guided label refinement to promote cross-domain knowledge transfer, and validates their effectiveness on three benchmarks.


\section*{Acknowledgements}
This paper was supported by National Key R\&D Program of China (No. 2021YFB3301503), and also supported by the National Natural Science Foundation of China (No. 11727801).

\bibliographystyle{ACM-Reference-Format}
\bibliography{reference}

\end{document}